\documentclass[11pt,a4paper]{article}
\usepackage[hyperref]{acl2018}
\usepackage{times}
\usepackage{latexsym}
\usepackage{graphicx}
\graphicspath{ {pic/} }
\usepackage{amsmath,amssymb}
\usepackage{eucal}
\usepackage{algorithm}  
\usepackage{algorithmic}

\usepackage{url}
\usepackage{multirow}

\aclfinalcopy % Uncomment this line for the final submission
\title{Deep Reinforcement Learning for On-line Dialogue State Tracking}

\author{Zhi Chen, Lu Chen, Xiang Zhou and Kai Yu \\
  Key Lab. of Shanghai Education Commission for Intelligent Interaction and Cognitive Eng. \\
  SpeechLab, Department of Computer Science and Engineering \\
  Brain Science and Technology Research Center \\
  Shanghai Jiao Tong University, Shanghai, China \\
  {\tt \{zhenchi713,chenlusz,owenzx,kai.yu\}@sjtu.edu.cn} }

\date{}

\begin{document}
\maketitle
\begin{abstract}
Dialogue state tracking (DST) is a crucial module in dialogue management. It is usually cast as a supervised training problem, which is not convenient for on-line optimization. In this paper, a novel companion teaching based deep reinforcement learning (DRL) framework for on-line DST optimization is proposed. To the best of our knowledge, this is the first effort to optimize the DST module within DRL framework for on-line task-oriented spoken dialogue systems. In addition, dialogue policy can be further jointly updated. Experiments show that on-line DST optimization can effectively improve the dialogue manager performance while keeping the flexibility of using predefined policy. Joint training of both DST and policy can further improve the performance.
\end{abstract}

\section{Introduction}
\begin{figure}[htbp!]
\centering
\vspace{-0.3cm}
\includegraphics[width=0.47\textwidth]{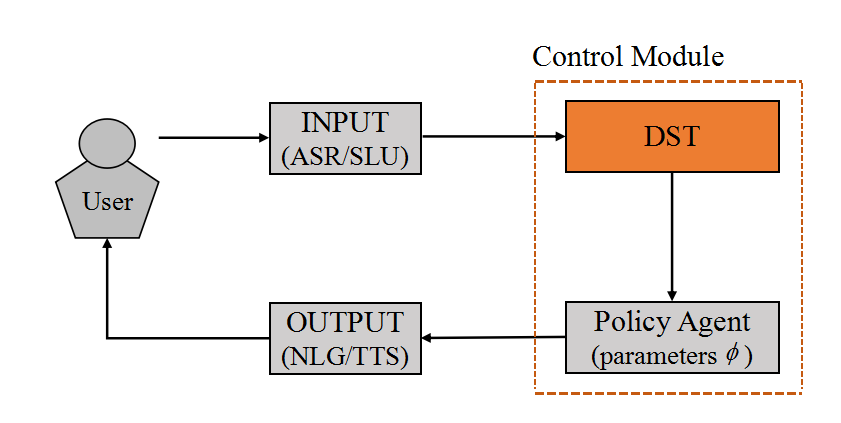}
\vspace{-0.3cm}
\caption{Spoken dialogue system. }
\vspace{-0.5cm}
\label{fig:SDS}
\end{figure}

A task-oriented spoken dialogue system usually consists of three modules: input,output and control, shown in Fig.\ref{fig:SDS}. The input module which consists of automatic speech recognition (ASR) and spoken language understanding (SLU) extracts semantic-level user dialogue actions from user speech signal. The control module (referred to as dialogue management) has two missions. One is to maintain dialogue state, an encoding of the machine's understanding about the conversation. Once the information from the input module is received, the dialogue state is updated by dialogue state tracking (DST). The other is to choose a semantic-level machine dialogue action to response the user, which is called dialogue decision policy. 
The output consists of natural language generation (NLG) and text-to-speech (TTS) synthesis, which convert dialogue action to audio. Dialogue management is an important part of a dialogue system. Nevertheless, there are inevitable ASR and SLU errors which make it hard to track true dialogue state and make decision. In recent statistical dialogue system, the distribution of dialogue state, i.e. {\em belief state}, is tracked. A well-founded theory for belief tracking and decision making is offered by partially observable Markov Decision Process (POMDP)~\cite{Kaelbling:1998ew} framework.

Previous DST algorithms can be divided into three families: hand-crafted rules~\cite{wang2013simple,sun2014generalized}, generative models~\cite{bui2009tractable,young2010hidden}, and discriminative models~\cite{sun2014sjtu,henderson2014word}. Recently, since the Dialog State Tracking Challenges (DSTCs) have provided labelled dialog state tracking data and a common evaluation framework and test-bed, a variety of machine learning methods for DST have been proposed. These methods rely strictly on set of labelled off-line data. Since the labelled data are off-line, the learning process of these supervised learning methods is independent on the dialogue policy module. The key issues of these supervised learning methods are poor generalization and over-tuning. Due to the lack of labels, these approaches can not be easily used for on-line update of DST. 

This work marks first step towards employing the deep reinforcement learning (DRL) method into dialogue state tracking (DST) module. The performance of the DST module is optimized during the conversation between the user and the dialogue system. We call the DRL-based DST module as the tracking agent. In order to bound the search space of the tracking agent, we propose a companion teaching framework \cite{chen2017line}. Furthermore, under this framework, we can train tracking agent and dialogue policy agent jointly with respective deep reinforcement learning (DRL) algorithms in order to make these two agents adaptive to each other.
% And there are two main types of DST systems in the current dialogue system. One is the semantic-based dialogue state tracking system, and the other is the text-based dialogue state tracking system that implicitly or explicitly removes spoken language understanding (SLU) module. In this paper, we explain the proposed tracking agent framework based on the semantic-based dialogue state tracking (DST) system.

The paper has two main contributions:
\begin{itemize}
\vspace{-0.3cm}
\item The paper provides a flexible companion teaching framework which makes the DST be able to be optimized in the on-line dialogue system.
\vspace{-0.3cm}
\item We can jointly train DST agent and dialogue policy agent with different reinforcement learning algorithms. 
\end{itemize}

The rest of the paper is organized as follows. Section \ref{sec:rw} gives an overview of related work. In Section \ref{sec:DSTRL}, the framework of on-line DST are presented. The implementation detail is represented in Section \ref{sec:Implementation}. In Section \ref{sec:joint}, the joint training process is introduced. Section \ref{sec:experiments} presents experiments conducted to evaluate the proposed framework, followed by the conclusion in Section \ref{sec:conclusion}.

\section{Related Work}
\label{sec:rw}
Recent mainstream studies on dialogue state tracking are discriminative statistical methods. Some of the approaches encode dialogue history in features to learn a simple classifier. \citeauthor{henderson2013deep} applies a deep neural network as a classifier. \citeauthor{williams2014web} proposed a ranking algorithm to construct conjunctions of features. The others of the approaches model dialogue as a sequential process \cite{williams2016dialog}, such as conditional random field (CRF)~\cite{lee2013structured} and recurrent neural network (RNN)~\cite{henderson2014word}. All of these approaches need massive labelled in-domain data for training, they are belong to off-line and static methods.

In contrast to dialogue state tracking, the dialogue policy in task-oriented SDS has long been trained using deep reinforcement learning (DRL) which includes value function approximation methods, like deep Q-network (DQN) \cite{cuayahuitl-nipsworkship2015,zhao-sigdial2016,lipton-arxiv2016,Fatemi:2016tr,chen2017line,chang-emnlp2017}, and policy gradient methods, e.g. REINFORCE \cite{su2017sample,williams2017hybrid}, advantage actor-critic (A2C) \cite{Fatemi:2016tr}. \cite{mnih-nature2015} under POMDP framework. In our experiments, the dialogue policies of our provided systems are optimized by DQN. Our proposed framework is also inspired by the success of the companion teaching methods \cite{chen2017line} in the dialogue policy.

In this work, we propose a companion teaching framework to generate the tracker from the on-line dialogue system. But the space of the belief state is continuous, it is difficult to be optimized by normal RL algorithms. \citeauthor{hausknecht2015deep} provided an efficient method to extend deep reinforcement learning to the class of parameterized action space MDPs and extend the deep deterministic policy gradient (DDPG) algorithm~\cite{lillicrap2015continuous} for bounding the action space gradients suggested by the critic. This method greatly reduces the difficulty of the exploration.

The closest method is the Natural Actor and Belief
Critic (NABC) algorithm~\cite{jurvcivcek2012reinforcement} which jointly optimizes both the tracker and the policy parameters. However, the tracker in ~\cite{jurvcivcek2012reinforcement} uses a dynamic Bayesian Network to represent the dialogue state. In our work, the dialogue management system which includes the DST and the dialogue policy is purely statistical. Recently,~\citeauthor{liu2017end} proposed an end-to-end dialogue management system, which directly connects the dialogue state tracking module and the dialogue decision policy module with the reinforcement learning method. There are many problems with this approach in the practical dialogue system, where end-to-end dialogue systems are not as flexible and scalable as the modular dialogue system.

% tracking policy -. tracking agent (input output)
% policy -. policy agent

\section{On-line DST via Interaction}
\label{sec:DSTRL}
\begin{figure}[htbp!]
\centering
\vspace{-0.3cm}
\includegraphics[width=0.5\textwidth]{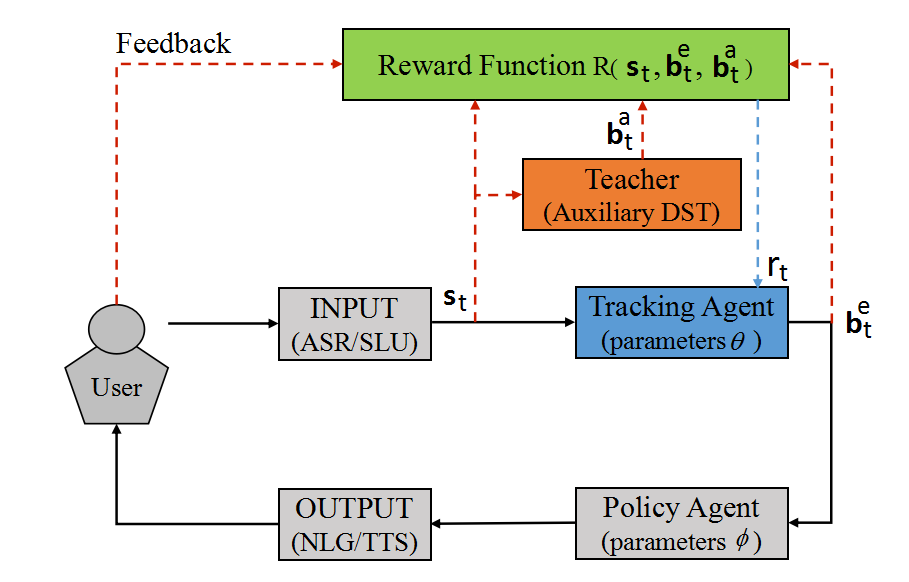}
\vspace{-0.4cm}
\caption{The companion teaching framework for the on-line DST. }
\vspace{-0.3cm}
\label{fig:DSTRL}
\end{figure}
Discriminative machine learning approaches are now the state-of-the-art in DST. However, these methods have several limitations. Firstly, they are supervised learning (SL) approaches which require massive off-line data annotation. This is not only expensive but also infeasible for on-line learning. Secondly, given limited labelled data, over-tuning may easily happen for SL approaches, which leads to poor generalization. Thirdly, since SL-based DST approaches are independent of dialogue policy, the DST module can't dynamically adapt to the habit of the user. These limitations prohibit DST module from on-line update. To address this problem, we propose a deep reinforcement learning (DRL) framework for DST optimization via on-line interaction. 

Reinforcement learning (RL) has been popular for updating the dialogue policy module in a task-oriented dialogue system for a long time. However, except for a few joint learning model for both DST and policy, RL has not been used specifically for the DST module. In this paper, under the RL framework, we regard the DST as an agent, referred to as {\em tracking agent}, and the other parts of the dialogue system as the environment. To our best knowledge, this is the first attempt to employ the reinforcement learning framework specifically for on-line DST optimization.  

Different from the policy agent, the decision (belief state) made by the tracking agent is continuous. Hence,  in this paper, DST is cast as a continuous control problem which has a similar challenge as robot control. There are several advanced algorithms to tackle the continuous control problems, e.g. DDPG algorithm. However, since the continuous belief state is both continuous and high-dimensional, the straightforward application of existing RL algorithms do not work well. 
In this paper, we borrow the {\em companion teaching} idea \cite{chen2017line} to construct a novel RL framework for DST. Here, an auxiliary well-trained tracker, e.g. a traditional tracker trained off-line, is used as the {\em teacher} to guide the optimizing process of the actual DST agent (the {\em student}) to avoid over-tuning and achieve robust and fast convergence. 

The companion teaching RL-DST framework is shown in Fig.\ref{fig:DSTRL}, where ${\bf b}^{\tt a}$ is the auxiliary belief state produced by the auxiliary DST model and ${\bf b}^{\tt e}$ is the exploration belief state produced by the tracking agent. The difference between ${\bf b}^{\tt a}$ and ${\bf b}^{\tt e}$ will be fed into the reward signal to significantly reduce the search space of the tracking agent. 

It is also worth comparing the proposed framework with end-to-end dialogue systems which can also support on-line update. Firstly, the modular structure of the RL-DST framework allows more flexible and interpretable dialogue management models to be used. For example, interpretable dialogue policy, such as rule-based policy, can be easily used with arbitrary DST models. This flexibility is practically very useful. Secondly, due to the use of a teacher DST model, the optimizing process of the tracking agent requires few dialogue data and the training is more robust. 

%In the next three subsections, we will introduce the RL components of the tracking agent and how the traditional DST bounds the search space of the tracking agent in detail. 
\subsection{Input \& Output}
To avoid confusion with the concepts of the policy agent, we replace {\em input} and {\em output} for {\em state} and {\em action} of the tracking agent respectively. In this work, only semantic-level dialogue manager is considered. Thus, the input is semantic features of each slot, which are extracted from system action, spoken language understanding (SLU) output and context from the previous turn. The output of the tracking agent is belief state of the corresponding slot at the current turn. In contrast to the system action of the policy agent, the output of the tracking agent, i.e. belief state, is continuous. In this paper, the input of the tracking agent is represented as $\textbf{s}$ and the output as $\textbf{b}^{\tt e}$.

\subsection{Tracking Policy}
The {\em tracking policy} denotes a mapping function between $\textbf{s}$ and $\textbf{b}^{\tt e}$ which aims to maximize the expected accumulated reward. Since the search space of the tracking agent is continuous, deterministic reinforcement learning algorithms, such as DDPG algorithm, is used to optimize the tracking policy as in the robotic control problem~\cite{lillicrap2015continuous,gu2017deep}.

\subsection{Reward Signal}
\label{ssec:reward}

Dialogue system reward is usually defined as a combination of \textbf{turn penalty} and \textbf{success reward} \cite{cuayahuitl-nipsworkship2015,zhao-sigdial2016}. The policy agent can be effectively optimized using the two reward signals. However, for the tracking agent, due to the large search space caused by the continuous output, the two signals are not sufficient to achieve fast and robust convergence. To address this problem, we design another \textbf{basic score} reward signal to constrain the search space of the tracking agent. Therefore, the overall reward of the tracking agent consists of three kinds of signals:

\textbf{Turn Penalty}, denoted as $r^{\tt tp}$, is a  negative constant value to penalize long dialogues. The assumption here is that shorter dialogue is better.

\textbf{Success Reward}, denoted as $r^{\tt sr}$, is a delayed reward for the whole dialogue at the last turn. When the conversation between the user and the machine is over, the user gives an evaluation value to judge the performance of the dialogue system. If the whole conversation has not achieved the user's goal, the success reward will be zero. Otherwise, success reward will be a positive constant value. 

\textbf{Basic Score}, denoted as $r^{\tt bs}$, is used to reduce the search space for tracking agent. As shown in Fig.\ref{fig:DSTRL}, an auxiliary DST is used. We use the auxiliary belief state $\textbf{b}^{\tt a}$ to guide the exploration of the tracking agent. If the exploration belief state $\textbf{b}^{\tt e}$ is far away from the auxiliary belief state, a penalty is given as in equation (\ref{eq:basic_score}). Thus, basic score is inversely proportional to the distance between auxiliary belief state and exploration belief state.
\begin{equation}
r^{\tt bs} = -\alpha||\textbf{b}^{\tt e}-\textbf{b}^{\tt a}||_2 \nonumber
\label{eq:basic_score}
\end{equation}
where $||\cdot||_2$ is L2 distance and $\alpha \geq 0$ is referred to as \textbf{{\em trust factor}}. With larger $\alpha$, performance of the tracking agent is closer to the auxiliary DST model. 

% \textbf{Exploration Reward} We know that standard action isn't the best action. In the case, exploration actions make the whole conversation accept by user. Intuitively, we should give reward to these exploration action. This mechanism is in order to simulate the tracking agent to explore bravely. Opposite to standard scoring strategy, exploration reward is proportional to the distance between standard action and exploration action. 
% \begin{equation}
% r_4=\left\{  
%              \begin{array}{ll}  
%              \lambda_{2}\sum_{t=1}^{T}(a_{t}^{\tt e}-a_{t}^{s})^2, &  success=True\\   
%              0, &  success=False  
%              \end{array}  
% \right. 
% \label{eq:exploration_reward}
% \end{equation}
% where $\lambda_{2} \geq 0$. 

% This is also a delayed reward which is given at the last turn. If the whole conversation is failed, exploration reward will be set to zero. 

In the middle of a conversation, the immediate reward of the exploration belief state is $r^{\tt tp} + r^{\tt bs}$, the immediate reward of the last turn is $r^{\tt sr}$.

\section{Implementation Detail}
\label{sec:Implementation}
In the companion teaching RL-DST framework, the auxiliary DST can make use of arbitrary well trained DST model and the tracking agent can be optimized by any deterministic reinforcement learning algorithm. In this section, we will introduce the dialogue tasks as well as the specific algorithm implementations, though the actual algorithms are not constrained to the below choices. Note that, the tracking agent, i.e. the DST to be optimized, takes a form of deep neural network in this paper.   

In this work, we evaluate the proposed framework on the task-oriented dialogue systems in the restaurant/tourism domain in DSTC2/3~\cite{henderson2014second,henderson2014third}. These systems are $slot-based$ dialogue systems. There are three slot types: \textbf{goal constraint}, \textbf{request slots} and \textbf{search method}. The \textbf{goal constraint}s are constraints of the information/restaurant which the user is looking for. The \textbf{search methods} describe the way the user is trying to interact with the system. The \textbf{request slots} are demands which the user has requested. The three different types of slots have different influences on the dialogue performance. Therefore, we use multiple tracking agents, each agent per type, to represent dialogue tracking policy instead of only one overall tracking agent. Each agent has its own input and output. The final overall output is simply the concatenation of the outputs from all agents. 

\subsection{Auxiliary Polynomial Tracker}
\label{subsec:cmbp}
In this paper, a polynomial tracker is used as the auxiliary DST. It is also referred to as {\em Constrained Markov Bayesian Polynomial} (CMBP)~\cite{yu2015constrained} which is a hybrid model combining both data-driven and rule-based models. CMBP has small number of parameters and good generalization ability. In CMBP, the belief state at current turn is assumed to be dependent on the observations of the current turn and the belief state of the previous turn. A general form of CMBP is shown as below:

\begin{small} 
\begin{align}
b_{t+1}(v)=\CMcal{P}&(P_{t+1}^{+}(v),P_{t+1}^{-}(v),\tilde{P}_{t+1}^{+}(v),\tilde{P}_{t+1}^{-}(v),b_t^r,b_t(v)) \nonumber \\
&{\tt s.t. constraints}
\label{eq:CMBP_tracking}
\end{align}
\end{small}
where $\CMcal{P}(\cdot)$ is a polynomial function, $b_{t+1}(v)$ which denotes the probability of a specific slot taking value $v$ at the $(t+1)^{th}$ turn is a scalar value and constraints include probabilistic constraints, intuition constraints and regularization constraints. And there are six probabilistic features for each $v$ defined as below
\begin{itemize}
\item $P_{t+1}^{+}(v)$: sum of scores of SLU hypotheses informing or affirming value $v$ at turn $t+1$ 
\item $P_{t+1}^{-}(v)$: sum of scores of SLU hypotheses denying or negating value $v$ at turn $t+1$  
\item $\tilde{P}_{t+1}^{+}(v) = \sum_{v^{\prime} \notin \{v,None\}}P_{t+1}^{+}(v^{\prime})$
\item $\tilde{P}_{t+1}^{-}(v) = \sum_{v^{\prime} \notin \{v,None\}}P_{t+1}^{-}(v^{\prime})$
\item $b_t^r$: probability of the value being `None' (the value not
mentioned) at turn $t$
\item $b_t(v)$: belief of ``the value being $v$ at turn $t$''
\end{itemize}

\begin{figure}[htbp!]
\centering
\vspace{-0.3cm}
\includegraphics[width=0.5\textwidth]{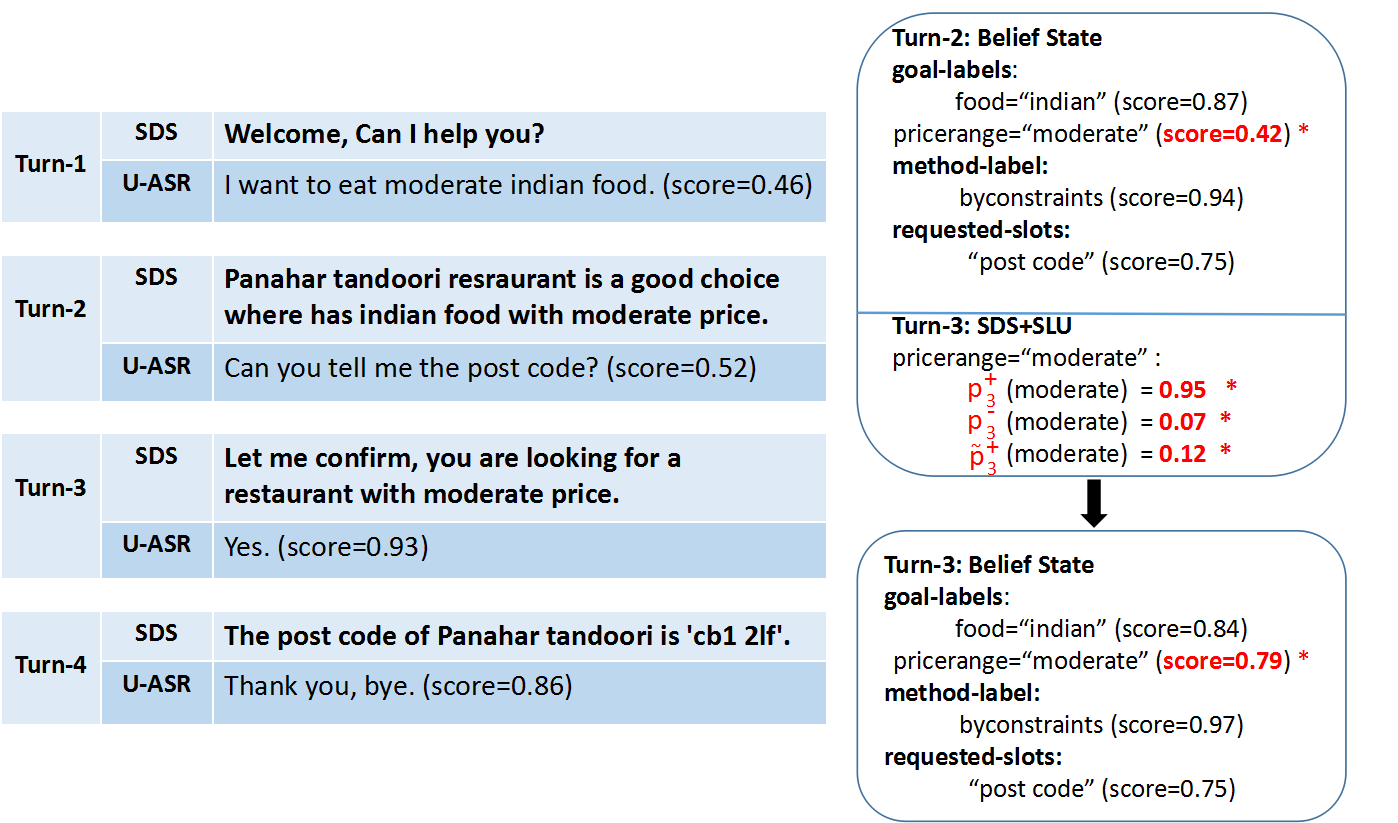}
\vspace{-0.4cm}
\caption{Example with polynomial dialogue state tracking.}
\label{fig:dialogue_example}
\end{figure}

In this paper, polynomial order is 3. The coefficients of polynomial $\CMcal{P}(\cdot)$ are optimized by the off-line pre-collected training data. Each slot type in DSTC2/3 (goal, request, method) has its own polynomial model, represented by $\CMcal{P}_g(\cdot)$,  $\CMcal{P}_r(\cdot)$ and  $\CMcal{P}_m(\cdot)$ respectively. The belief state of different slot-value pairs within the same slot type is updated by the same polynomial. For example, in our work, we set $\CMcal{P}_g(\cdot)$ as:
\begin{align*}
b_{t+1}(v)=&(b_t(v)+P_{t+1}^{+}(v)*(1-b_t(v)))* \\ 
&(1-P_{t+1}^{-}(v)-\tilde{P}_{t+1}^{+}(v))
\end{align*}
An example of updating belief state of the slot {\em pricerange} using polynomial tracker is shown in Fig.\ref{fig:dialogue_example}.

\subsection{Tracking Agents}
\begin{figure}[htbp!]
\centering
\includegraphics[width=0.47\textwidth]{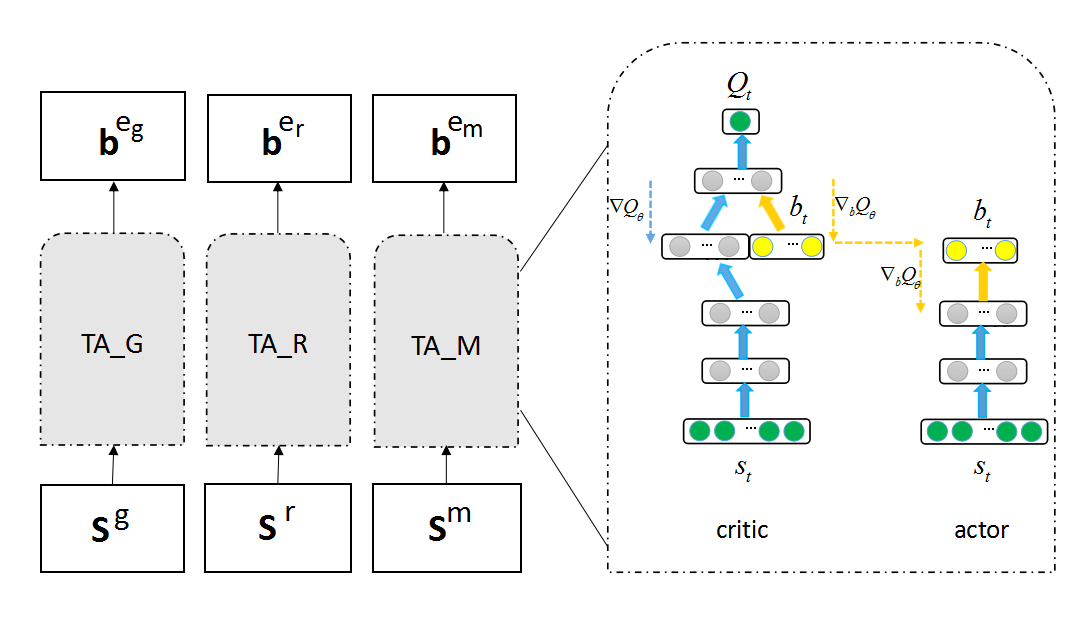}
\vspace{-0.3cm}
\caption{Multi-agent Tracking in DSTC2/3. }
\vspace{-0.4cm}
\label{fig:MAT_DDPG}
\end{figure}

Three types of the slots (goal, request, method) in DSTC2/3 are not to affect each other. Therefore, the DST tracking agent can be decomposed into three independent tracking agents for DSTC2/3 tasks, represented by \textbf{TA\_G}, \textbf{TA\_R} and \textbf{TA\_M} in Fig.\ref{fig:MAT_DDPG}. These tracking agents have individual RL components as described in section \ref{sec:DSTRL}. The three tracking agents correspond to the auxiliary DST trackers $\CMcal{P}_g(\cdot)$,  $\CMcal{P}_r(\cdot)$ and  $\CMcal{P}_m(\cdot)$ respectively. Note that the forms of the DST tracking agents are deep neural networks instead of polynomials.

In this paper, the input of each tracking agent which represents as $\textbf{s}^g$, $\textbf{s}^r$ and $\textbf{s}^m$ is consistent with the input of polynomial tracker represented in equation (\ref{eq:CMBP_tracking}) where each slot is represented by six probabilistic features. The output of each tracking agent represented by $\textbf{b}^{\tt e_g}$, $\textbf{b}^{\tt e_r}$ and $\textbf{b}^{\tt e_m}$ in Fig.\ref{fig:MAT_DDPG} is belief state of corresponding slots at the next turn. In this work, we adopt deep deterministic policy gradient (DDPG) algorithm to optimize these three tracking agents. The flexibility of our framework is that we can optimize the selective parts of DST module, the other parts of belief state can still be produced by the auxiliary polynomial DST. We can also figure out which parts of DST module have the bigger effect on the dialogue performance. 

\subsection{DDPG for Tracking Policy}
In order to optimize three tracking agents in Fig.\ref{fig:MAT_DDPG} which have continuous and high dimensional output spaces, we use DDPG algorithm~\cite{lillicrap2015continuous} which is an actor-critic, model-free algorithm based on the deterministic policy gradient that can operate over continuous action spaces. This algorithm combines the actor-critic approach with insights from the DQN algorithm which has a replay buffer and adopts the soft-update strategy. 

During training of the three agents, there are three experience memories for each tracking agent respectively. The format of the data in memories is  $(\textbf{s}_t, \textbf{b}^{\tt e}_t, r_t, \textbf{s}_{t+1})$. $\textbf{s}_t$ is slot feature vector and $\textbf{b}^{{\tt e}}_t$ is the exploration belief state of corresponding slots. The immediate reward $r_t$ is produced by reward function $R(\textbf{s}_t,\textbf{b}^{{\tt e}}_t, \textbf{b}^{\tt a}_t)$ at each turn, presented at Section \ref{ssec:reward}.

The DDPG algorithm uses the deterministic policy gradient (DPG) method to update the deep neural networks. 
There are two functions in DDPG algorithm: the actor policy function $\pi(\textbf{s}_t|\theta)$ deterministically maps the input to output, and the critic function $Q(\textbf{s}_t,\textbf{b}^{{\tt e}}_t|\beta)$ is learned using the Bellman equation as in Q-learning which aims to minimize the following loss function,
\begin{equation}
L(\beta) = {\mathbb E}_{\textbf{s}_t,\textbf{b}^{{\tt e}}_t,r_t}[(Q(\textbf{s}_t,\textbf{b}^{{\tt e}}_t|\beta)-y_t)^2]
\label{eq:critic_loss}
\end{equation}
where $y_t = r_t +\lambda Q(\textbf{s}_{t+1},\pi(\textbf{s}_{t+1}|\beta))$, $r_t)$ is immediate reward at $t^{th}$ turn  and $\lambda \in [0,1]$ is discount factor.

The target of the actor policy is to maximize the cumulative discounted reward from the start state, denoted by the performance objective $J(\pi)={\mathbb E}[\sum_{t=1}^{T}\gamma^{t-1}r_t|\pi]$. \citeauthor{silver2014deterministic} proved that the following equation is the actor policy gradient, the gradient of the policy' s performance:
\begin{align}
\nabla_{\theta}J \approx {\mathbb E}_{\textbf{s}_t}[\nabla_{\textbf{b}^{{\tt e}}}Q(\textbf{s},\textbf{b}^{{\tt e}}|\beta)|_{\textbf{s}=\textbf{s}_t,\textbf{b}^{{\tt e}}=\pi(\textbf{s}_t)} 
\nabla_{\theta}\pi(\textbf{s}|\theta)|_{\textbf{s}=\textbf{s}_t}]
\label{eq:DDPG}
\end{align}
where $\nabla_{\textbf{b}^{{\tt e}}}Q(\textbf{s},\textbf{b}^{{\tt e}}|\beta)$ denotes the gradient of the critic with respect to actions and $\nabla_{\theta}\pi(\textbf{s}|\theta)$ is a Jacobian matrix such that each column is the gradient $\nabla_{\theta}[\pi(\textbf{s}|\theta)]_d$ of the $d^{th}$ action dimension of the policy with respect to the policy parameters $\theta$. The implementation details of the DDPG algorithm are provided in~\cite{lillicrap2015continuous}.

\section{Joint Training Process}
\label{sec:joint}
In Section \ref{sec:Implementation}, we discuss the implementation details of the on-line DST in DSTC2/3 cases. During the learning process of the tracking agent, the dialogue policy is fixed and the tracker keeps changing. Since the DST is part of the environment for the dialogue policy agent, when the tracking agent is optimized, the environment of the dialogue policy agent is also changed. Thus, we can choose to further optimize the dialogue policy in order to get even more improved dialogue system performance. This is referred to as {\em joint training} of DST and policy. The process of joint training consists of four phases: the pre-training of the dialogue policy agent, the pre-training of the tracking agent, the training of the dialogue policy agent and the training of the tracking agent. The details of the joint training shows in algorithm \ref{code:jointTrain}.

\begin{algorithm}[h]  
\caption{The process of joint training}  
\begin{algorithmic}[1]
\STATE Initialize dialogue policy $Q(\phi)$, \textbf{TA\_G} tracking agent $Q(\beta^g), \pi(\theta^g)$, \textbf{TA\_R} tracking agent $Q(\beta^r), \pi(\theta^r)$, \textbf{TA\_M} tracking agent $Q(\beta^m), \pi(\theta^m)$ \\
{\em $//$ pre-train dialogue policy agent}
\STATE Set polynomial method as the tracker of the system 
\FOR{$episode=1:N_1$}  
\STATE Update dialogue policy using DQN algorithm  
\ENDFOR \\
{\em $//$ pre-train tracking agents}
\FOR{$episode=1:N_2$}  
\STATE Update the actors $\pi(\theta^g)$, $\pi(\theta^r)$, $\pi(\theta^m)$ of the tracking agents by minimizing mean squared error with the output of polynomial tracker $\textbf{b}^{\tt a_g}$, $\textbf{b}^{\tt a_r}$ and $\textbf{b}^{\tt a_m}$. 
\ENDFOR   \\
{\em $//$ optimize tracking agents}
\STATE Set multi-tracking agent as the tracker of the system 
\FOR{$episode=1:N_3$}
\STATE Update the critics $Q(\beta^g)$, $Q(\beta^r)$, $Q(\beta^m)$ of the multi-tracking agent by minimizing equation (\ref{eq:critic_loss})
\STATE Update the actors $\pi(\theta^g)$, $\pi(\theta^r)$, $\pi(\theta^m)$ of the multi-tracking agent by equation (\ref{eq:DDPG})
\ENDFOR \\
{\em $//$ optimize dialogue policy agent}
\FOR{$episode=1:N_4$}  
\STATE Update dialogue policy using DQN algorithm  
\ENDFOR  
\end{algorithmic}  
\label{code:jointTrain}
\end{algorithm}  

\section{Experiments}
\label{sec:experiments}
\begin{figure*}[htbp!]
\centering
\vspace{-0.3cm}
\includegraphics[width=\textwidth]{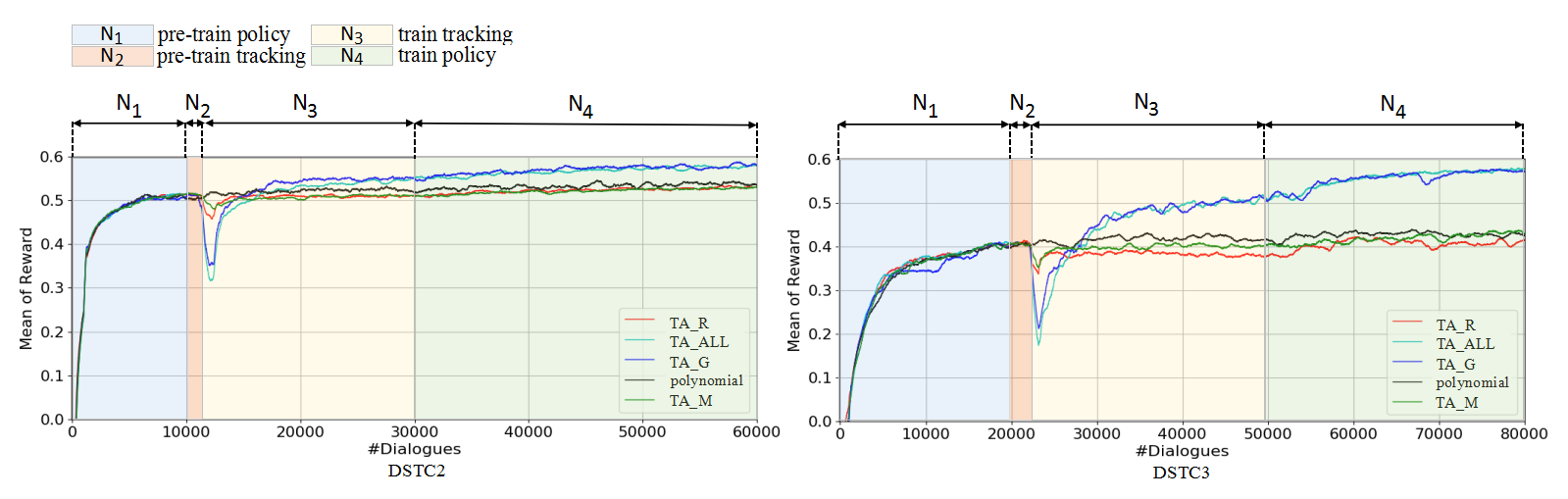}
\vspace{-1.0cm}
\caption{The learning curves of joint training dialogue systems and baseline system in DSTC2 (left) and DSTC3 (right).}
\vspace{-0.3cm}
\label{fig:reward}
\end{figure*}
Two objectives are set for the experiments: (1) Verifying the performance of the SDS with the optimized on-line DST. (2) Verifying the performance of the dialogue system which jointly train the DST and the dialogue policy.

\subsection{Dataset}
The proposed framework is evaluated on domains of DSTC2/3 ~\cite{henderson2014second,henderson2014third}. In DSTC2, there are 8 requestable slots and 4 informable slots. In DSTC3, there are 12 requestable slot and 8 informable slots. Therefore the task in DSTC3 is more complex. Furthermore, the semantic error rate in  DSTC3 is higher than the semantic error rate in DSTC2.

Based on the datasets in DSTC2/3, an agenda-based user simulator  \cite{Schatzmann:2007em} with error model \cite{schatzmann2007error} was implemented to emulate the behaviour of the human user and errors from the input module.

\subsection{Systems}
In our experiments, six spoken dialogue systems with different DST models were compared: 
\begin{itemize}
\item \textbf{Polynomial} is the baseline system. The {\em polynomial} DST as described in section \ref{subsec:cmbp} is used. The corresponding policy agent is a two-layer DQN network with 128 nodes per layer.  
\vspace{-0.2cm}
\item \textbf{TA\_G} is a DST tracking agent as in Fig.\ref{fig:MAT_DDPG}. It only estimates the belief state of \textbf{goal constraint} and the other two parts of belief state are produced by the polynomial tracker. 
\vspace{-0.2cm}
\item \textbf{TA\_R} is a DST tracking agent as in Fig.\ref{fig:MAT_DDPG}. It only estimates the belief state of \textbf{request slots} and the other two parts of belief state are produced by the polynomial tracker.
\vspace{-0.2cm}
\item \textbf{TA\_M} is a DST tracking agent as in Fig.\ref{fig:MAT_DDPG}. It only estimates the belief state of \textbf{search method} and the other two parts of belief state are produced by the polynomial tracker.
\vspace{-0.2cm}
\item \textbf{TA\_ALL} is a DST tracking agent as in Fig.\ref{fig:MAT_DDPG}. Here, the whole belief state is directly produced by the above three tracking agents.
\vspace{-0.2cm}
\item \textbf{TA\_noteaching} is similar to \textbf{TA\_ALL} except the \textbf{basic score} reward signal is not used. This is equivalent to directly on-line directly train a neural network DST tracker.
\end{itemize}

In traditional supervised-learning based DST approaches, metrics such as accuracy or L2 norm are used for evaluation. However, on-line DST optimization does not require semantic annotation and the optimization objective is to improve dialogue performance. Hence, in this paper, metrics for dialogue performance are employed to evaluate on-line DST performances. 
There are two metrics used for evaluating the dialogue system performance: average length and success rate. For the reward in section \ref{ssec:reward}, at each turn, the turn penalty is $-0.05$ and the dialogue success reward is $1$. The summation of the two rewards are used for evaluation, hence, the reward in below experiment tables are between 0 and 1. The trust factors of the basic score in {\bf TA\_G}, {\bf TA\_R} and {\bf TA\_M} tracking agents are $0.2$, $0.2$, $4$ in DSTC2 and $0.07$, $0.07$, $4$ in DSTC3. For each set-up, the moving reward and dialogue success rate are recorded with a window size of 1000. The final results are the average of 25 runs. 

\subsection{DRL-based DST Evaluation}
\label{sec:evaluation}
In this subsection, we evaluate the performances of the systems with five different on-line DST models ({\bf TA\_G}, {\bf TA\_R}, {\bf TA\_M}, {\bf TA\_ALL} and {\bf TA\_noteaching}). The dialogue policy agents of these five systems are optimized by DQN for $N_1$ (10000/20000 in DSTC2/3) episodes/dialogues with the same polynomial trackers. Next, these five systems start to train tracking agents. In the first $N_2$ (1000 in DSTC2/3) episodes, we pre-train {\em actor} part of DDPG with the output of polynomial tracker using mean squared error (MSE) in all tracking agents. After pre-training, tracking agents of these five systems are optimized by DDPG for $N_3$ (19000/29000 in DSTC2/3) episodes. In the {\em polynomial} SDS, the dialogue policy agent is optimized for $N_1+N_2+N_3$ episodes. 

In Fig.\ref{fig:reward}, after the tracking agents are optimized for almost 10000 episodes, the tracking agents in these four on-line DST systems achieve the convergence nearly in DSTC2/3. It demonstrates that the companion teaching framework for the on-line DST is efficient. In Table \ref{tab:TPolicy2} and Table \ref{tab:TPolicy3} \footnote{Because of the limitation of space, the table only shows the standard deviation of the reward in this work, the standard deviations of the success rate and the turn can be reflected by the standard deviation}, the tracking agents in the {\bf TA\_ALL} system and the {\bf TA\_G} system improve the performances of the SDS significantly in DSTC2 and DSTC3. The tracking agents in the {\bf TA\_ALL} and the {\bf TA\_G} learned a tracking policy which can track the goal of the user accurately. Thus, compared with the {\em polynomial} system, the length of dialogue in these two systems decrease sharply. The rewards in these two systems increase significantly. The performances of the {\bf TA\_R} system and the {\bf TA\_M} system are similar with {\em polynomial} system. We can conclude that \textbf{goal constraint} plays a more important role in dialogue state than \textbf{request slots} and \textbf{search method}. The {\bf TA\_noteaching} system crashed during the optimizing process of the tracking agents. It reflects the effectiveness of our proposed companion teaching framework.
\begin{table}%[]
\centering
\small
\begin{tabular}{| c | c | c | c | }
\hline
{\bf DST} & {\bf Success} & {\bf \#Turn} & {\bf Reward} \\
\hline
Polynomial &$0.769$ &$ 5.013$ &$ 0.519 \pm 0.016$  \\
\hline
\hline
TA\_ALL &$\textbf{0.775} $ &$ 4.474 $ &$ \textbf{0.551} \pm 0.018$  \\
TA\_G &$0.767$ &$ \textbf{4.375}$ &$0.548 \pm 0.020$  \\
TA\_R &$0.763$ &$ 5.057$ &$0.510 \pm 0.022$  \\
TA\_M &$0.765$ &$ 5.121 $ &$0.509 \pm 0.018$  \\
\hline
\hline
TA\_noteaching &$-$ &$ - $ &$ - $  \\
\hline
\end{tabular}
\caption{The performances of tracking agents in DSTC2. The symbol '-' means the dialogue system crashed.}
\vspace{-0.2cm}
\label{tab:TPolicy2}
\end{table}

\begin{table}%[]
\centering
\small
\begin{tabular}{| c | c | c | c | }
\hline
{\bf DST} & {\bf Success} & {\bf \#Turn} & {\bf Reward} \\
\hline
Polynomial &$\textbf{0.744}$ &$ 6.566 $ &$ 0.415 \pm 0.077$  \\
\hline
\hline
TA\_ALL &$ 0.713 $ &$ \textbf{4.117} $ &$ \textbf{0.507} \pm 0.083$  \\
TA\_G &$0.719$ &$ 4.290 $ &$\textbf{0.505} \pm 0.075$  \\
TA\_R &$0.701$ &$ 6.438 $ &$0.379 \pm 0.028$  \\
TA\_M &$0.731$ &$ 6.540 $ &$ 0.404 \pm 0.021$  \\
\hline
\hline
TA\_noteaching &$-$ &$ - $ &$ - $  \\
\hline
\end{tabular}
\caption{The performances of tracking agents in DSTC3. The symbol '-' means the dialogue system crashed.}
\vspace{-0.1cm}
\label{tab:TPolicy3}
\end{table}

\subsection{Joint Training Evaluation}
In this subsection, we evaluate the performances of the systems (except for the {\bf TA\_noteaching} system ) which jointly train dialogue policy agent and tracking agent. In the first $N_1+N_2+N_3$ episodes, training processes of five models have been mentioned in Section \ref{sec:evaluation}. As shown in Fig.\ref{fig:reward}, in the latter $N_4$ (30000 in DSTC2/3) episodes, four models which contain tracking agents stop optimizing corresponding tracking agents and start to optimize dialogue policy agent and the baseline system continues to train dialogue policy agent. In Fig.\ref{fig:reward}, we compare the above five systems and the final performances show in Table \ref{tab:joint_train} and Table \ref{tab:joint_train3}. Compared with the results of the optimized tracking agents in Table \ref{tab:TPolicy2} and Table \ref{tab:TPolicy3}, the success rates in the {\bf TA\_ALL} system and the {\bf TA\_G} system increase significantly. It demonstrates that the dialogue policies in the {\bf TA\_ALL} and the {\bf TA\_G} have adapted the optimized tracking agents respectively. 

%The task of DSTC3 is more complex than that of DSTC2. The result shows that the performances of the {\bf TA\_ALL} system and the {\bf TA\_G} system are boosted more significantly by joint training in DSTC3 than DSTC2. This illustrates that the system with the tracking agent is more adaptive than polynomial system.

Compared the results in DSTC3 with the results in DSTC2, we can find that the boost of performance in DSTC3 is larger than that in DSTC2. The reason is that the semantic error rate of SLU in DSTC3 is higher than that in DSTC2, therefore the belief state tracker plays a more important role in DSTC3. These results also indicate that our proposed DRL-based tracker is robust to the input errors of SDS.

\begin{table}%[]
\centering
\small
\begin{tabular}{| c | c | c | c | }
\hline
{\bf DST} & {\bf Success} & {\bf \#Turn} & {\bf Reward} \\
\hline
% POLY(dqn40000) &$0.779$ &$ 5.012$ &$ 0.529$  \\
Polynomial &$0.784$ &$ 4.995$ &$ 0.535 \pm 0.015$  \\
\hline
\hline
TA\_ALL &$\textbf{0.810}$ &$ 4.566 $ &$ \textbf{0.581} \pm 0.022$  \\
TA\_G &$0.805$ &$ \textbf{4.497} $ &$\textbf{0.580} \pm 0.015$  \\
TA\_R &$0.782$ &$ 5.052 $ &$ 0.530 \pm 0.014$  \\
TA\_M &$0.782$ &$ 5.051 $ &$0.530 \pm 0.020$  \\
\hline
\end{tabular}
\caption{The performances of joint training in DSTC2.}
\label{tab:joint_train}
\end{table}

\begin{table}%[]
\centering
\small
\begin{tabular}{| c | c | c | c | }
\hline
{\bf DST} & {\bf Success} & {\bf \#Turn} & {\bf Reward} \\
\hline
% POLY(dqn40000) &$0.779$ &$ 5.012$ &$ 0.529$  \\
Polynomial &$0.754$ &$ 6.580$ &$ 0.425 \pm 0.071$  \\
\hline
\hline
TA\_ALL &$0.795$ &$ \textbf{4.317} $ &$ \textbf{0.578} \pm 0.064$  \\
TA\_G &$\textbf{0.800}$ &$ 4.579 $ &$0.571 \pm 0.068$  \\
TA\_R &$0.747$ &$ 6.654 $ &$0.414 \pm 0.069$  \\
TA\_M &$0.759$ &$ 6.605 $ &$0.429 \pm 0.022$  \\
\hline
\end{tabular}
\caption{The performances of joint training in DSTC3.}
\label{tab:joint_train3}
\end{table}

\section{Conclusion}
\label{sec:conclusion}
This paper provides a DRL-based companion teaching framework to optimize the DST module of the dialogue system. Under this framework, the tracker can be learned during the conversations between the user and the SDS rather than produced by the off-line methods. We can also choose to jointly train dialogue policy agent and the tracking agent under this framework. The experiments showed that the proposed companion teaching framework for the on-line DST system achieved promising performances in DSTC2 and DSTC3.

% \section*{Acknowledgments}

% The acknowledgments should go immediately before the references.  Do not number the acknowledgments section ({\em i.e.}, use \verb|\section*| instead of \verb|\section|). Do not include this section when submitting your paper for review.

% % include your own bib file like this:
% %\bibliographystyle{acl}
% %\bibliography{acl2018}
\bibliography{acl2018}
\bibliographystyle{acl_natbib}

\end{document}